\documentclass[10pt,twocolumn,letterpaper]{article}

\usepackage[pagenumbers]{cvpr} % To force page numbers, e.g. for an arXiv version

\usepackage[dvipsnames]{xcolor}

\definecolor{cvprblue}{rgb}{0.21,0.49,0.74}
\usepackage[pagebackref,breaklinks,colorlinks,citecolor=cvprblue]{hyperref}

\usepackage{graphicx}
\usepackage{booktabs}
\usepackage{url}            % simple URL typesetting
\usepackage{booktabs}       % professional-quality tables
\usepackage{amsfonts}       % blackboard math symbols
\usepackage{nicefrac}       % compact symbols for 1/2, etc.
\usepackage{microtype}      % microtypography
\usepackage{xcolor}         % colors

\usepackage[utf8]{inputenc} % allow utf-8 input
\usepackage[T1]{fontenc}    % use 8-bit T1 fonts
\usepackage{algorithm}% http://ctan.org/pkg/algorithms
\usepackage{algorithmicx}% http://ctan.org/pkg/algorithms
\usepackage[noend]{algpseudocode}
\algrenewcommand\alglinenumber[1]{\tiny #1:}
\usepackage{fp}
\usepackage{wrapfig,lipsum,epstopdf}
\usepackage{lipsum}
\usepackage{rotating}
\usepackage{multirow}
\usepackage{pifont}
\newcommand{\cmark}{\ding{51}}
\newcommand{\xmark}{\ding{55}}
\usepackage{array, makecell} 
\usepackage{xcolor,colortbl}
\usepackage{comment}
\usepackage{kantlipsum}
\usepackage{tabularx}
\usepackage{amsmath}
\usepackage{diagbox}
\usepackage{float}
\usepackage{caption}
\usepackage{graphicx}
\usepackage{subcaption}
\usepackage{etoolbox}
\usepackage{enumitem}
\usepackage{adjustbox}
\usepackage{arydshln} % Required for dashed lines in tables
\setlist{topsep=0pt,noitemsep,topsep=0pt,parsep=0pt,partopsep=0pt}
\definecolor{bgreen}{rgb}{0.0, 0.5, 0.0}
\definecolor{deepskyblue}{rgb}{0.0, 0.75, 1.0}

\usepackage[accsupp]{axessibility}  % Improves PDF readability for those with disabilities.

\title{MiniGPT4-Video: Advancing Multimodal LLMs for Video Understanding with Interleaved Visual-Textual Tokens}

\author{
Kirolos Ataallah {$^{1}$} \quad
Xiaoqian Shen {$^{1}$} \quad
Eslam Abdelrahman {$^{1}$} \quad
Essam Sleiman {$^{2}$} \\
Deyao Zhu {$^{1}$}  \quad
Jian Ding {$^{1}$}  \quad
Mohamed Elhoseiny {$^{1}$} \\
$^{1}$ KAUST \quad
$^{2}$ Harvard University\\
}

\begin{document}
\maketitle

\begin{abstract}
This paper introduces MiniGPT4-Video, a multimodal Large Language Model (LLM) designed specifically for video understanding. The model is capable of processing both temporal visual and textual data, making it adept at understanding the complexities of videos.
Building upon the success of MiniGPT-v2, which excelled in translating visual features into the LLM space for single images and achieved impressive results on various image-text benchmarks, this paper extends the model's capabilities to process a sequence of frames, enabling it to comprehend videos.
MiniGPT4-video does not only consider visual content but also incorporates textual conversations, allowing the model to effectively answer queries involving both visual and text components. The proposed model outperforms existing state-of-the-art methods,  registering gains of 4.22\%, 1.13\%, 20.82\%, and 13.1\% on the MSVD, MSRVTT, TGIF, and TVQA benchmarks respectively (see Figure~\ref{fig_teaser}).
Our models and code have been made publicly available  \href{https://vision-cair.github.io/MiniGPT4-video/}{here}.

\end{abstract}    

\section{Introduction}
\label{sec:intro}

In recent years, Large Language Models (LLMs) research has witnessed remarkable advancements, with prominent models like GPT-4~\cite{achiam2023gpt}, Llama 2~\cite{llama2}, and Mistral~\cite{jiang2023mistral} showcasing unprecedented capabilities in processing and generating textual data. However, typical LLMs are inherently limited to text-centric tasks and do not naturally capture the multimodal nature of human interaction with the world. While some strides have been made towards integrating images into LLMs, exemplified by works such as MiniGPT and LLaVa\cite{zhu2023minigpt,minigptv2,llava}, the incorporation of temporal information from videos remains relatively underexplored and presents significant research challenges.

Unlike static images, videos present a temporal dimension, comprising sequences of frames, essential for understanding dynamic visual content alongside textual input. In this study, we endeavor to adapt LLMs to comprehend the temporal intricacies inherent in video sequences. Previous efforts, such as Video-ChatGPT~\cite{videochatgpt}, have relied on spatial and temporal pooling techniques for fusing information across frames. However, these approaches often suffer from information loss and may not fully exploit the temporal dynamics of video data. Another example is LLaMA-VID~\cite{llamavid}, which attempted to address the constraints of LLMs in processing long videos by representing each frame with only two tokens, resulting in significant information loss.

% ---------------------------------------------
\begin{figure}
    \centering
    \scalebox{0.9}{
        \includegraphics[width=1\linewidth]{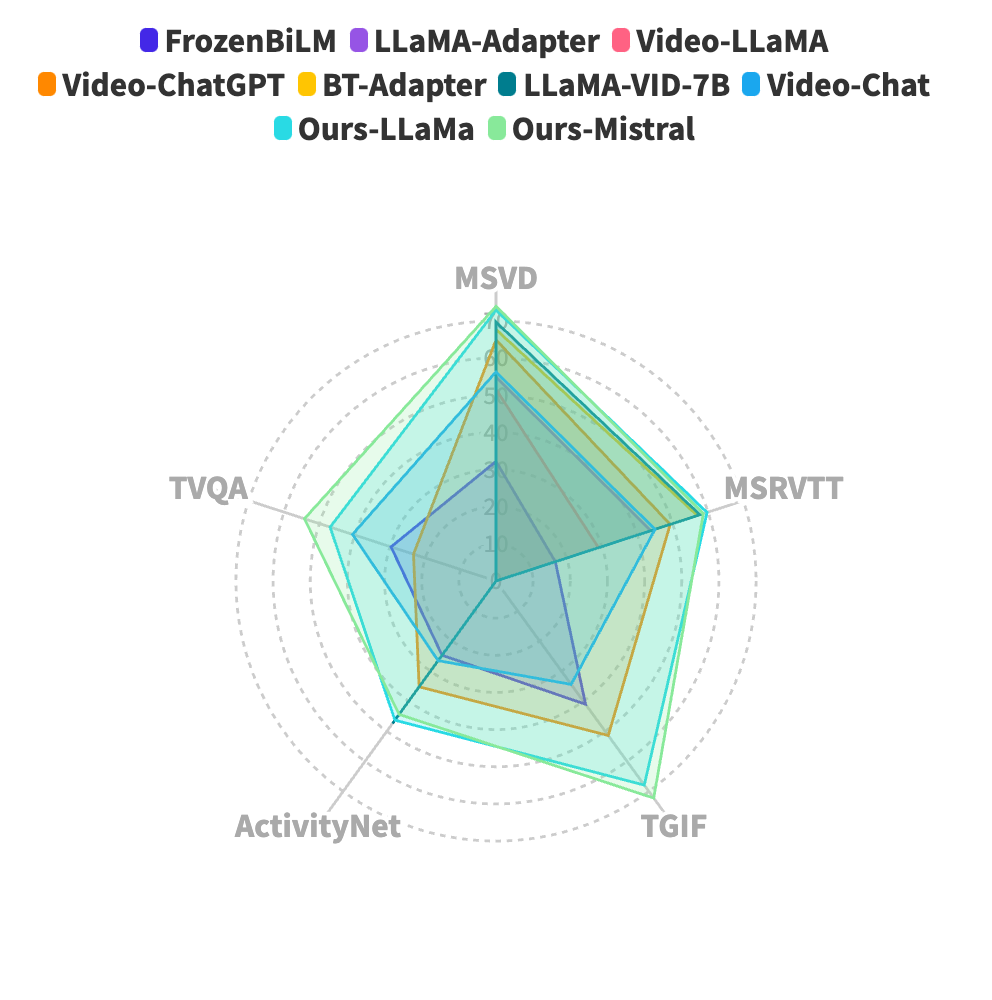}
        }
        \vspace{-9.0mm}
        \caption{Holostic qualitative results across five video benchmarks, namely, MSVD, MSRVTTm TGIF, ActivityNet, and TVQA.}
    \label{fig_teaser}
\end{figure}
% ---------------------------------------------

In contrast, our approach leverages the concatenation of every four adjacent visual tokens, effectively reducing the token count while mitigating information loss. As depicted in Figure~\ref{fig:main_fig}, we incorporate subtitles for each frame, allowing the representation of each frame as a combination of visual tokens extracted by the visual encoder\cite{sun2023eva} and text tokens derived from LLM tokenizers. Such an approach enables the LLM to comprehend the video content more comprehensively, facilitating responses to both visual and textual queries.

To validate the effectiveness of our proposed methodology, we conduct thorough evaluations across multiple benchmarks. These evaluations include assessments based on the Video-ChatGPT benchmark~\cite{videochatgpt}, which evaluates aspects such as information correctness, detail orientation, contextual understanding, temporal comprehension, and consistency in video understanding. Additionally, we employ zero-shot evaluation methodologies encompassing open-ended questions and multiple-choice formats. As shown in Figure \ref{fig_teaser}, the proposed MiniGPT4-Video outperforms existing state-of-the-art methods (7B) by notable margins of 4.22\%, 1.13\%, 20.82\%, and 13.1\% on the MSVD, MSRVTT, TGIF, and TVQA benchmarks, respectively.
% \elhoseiny{Do we have clarity on how we compare to these methods in terms of memory/computation cost? }
% \kirolos{No, we don't have these information for the other methods}

\section{Related work}
\label{sec:related_work}
%\subsection{Large Vision-Language Models}

\subsection{Large Vision-Language Models}
The advancements in NLP and LLMs have inspired a wide range of methods and models known as vision language models (VLM) for understanding cross-modalities~\cite{li2019visualbert,chen2020uniter,zhuge2021kaleido}. OpenAI's CLIP~\cite{clip} model aligns an image and language encoder on a large-scale dataset of image-text pairs with a contrastive loss, enabling it to perform multimodal-retrieval. More recent methods have harnessed the power of LLMs. For example, Flamingo~\cite{flamingo} leverages training on web-scraped image-text pairs to reveal the in-context learning capabilities of LVLMs. Similarly, BLIP-2~\cite{blip2}, which integrates off-the-shelf frozen pre-trained image encoders and large language models to bridge the modality gap with a Querying Transformer. 
There is a trend to use large-scale instruction-tuning datasets to fine-tune LVLMs. For example, LLaVA~\cite{llava} explores the instruction tuning of its vision language model on generated GPT-4 data, while MiniGPT-v2~\cite{minigptv2} and InstructBLIP~\cite{liu2023improved} use BLIP-2~\cite{blip2} to construct their datasets. As LLMs continue to achieve better results in image understanding, recent work has begun scaling these models to the more challenging video domain.
% gpt-4vision [citation], etc. all can do open-ended visual question-answering tasks.\\\\
% Blip-2 utilizes an off-the-shelf frozen visual encoder and LLM connected by a trained component called Q-former that learns to extract visual representation that is most informative of the text.
\subsection{LLM-Based Video Understanding}
% \jian{The term video-language models may not be very accurate. All the listed papers are LLM based video language models. Video-language models may also include previous works that does not use LLM. \jun{I agree with jian, you may consider Video-Language LLM}}\\
% \jian{Maybe add some video-language models that does not use LLM}
 % Video ChatGPT [citation], Video LLava [citation], Video Chat [citation], Llama Adapter [citation], Video Llama [citation], Llama Vid [citation] are some of these methods. Our approach, video-minigpt is better because
 % \jian{describe some topics of short videos}
Recently, vision-language models such as LLaVA~\cite{llava} have been extended to the video domain to process short videos 5 minutes on average or less, with similar capabilities such as visual question-answering and captioning. Video-LLaMA~\cite{video-llama} and VideoChat~\cite{li2024videochat} extend the BLIP-2~\cite{blip2} architecture for video embedding extraction and both employ two streams for audio and visual signals. Video-LLaMA employs a Video Q-Former and an Audio Q-Former for the two streams, while VideoChat has a video embedder and a perception toolkit for captioning, tags, etc.
% Video-LLaMA~\cite{video-llama} and VideoChat~\cite{li2024videochat} extend the BLIP-2~\cite{blip2} architecture for video embedding extraction. Video-LLaMA has two streams for both audio and visual inputs from video, employing a Video Q-Former for analyzing temporal image sequences and an Audio Q-Former for audio signals.  Similarly, VideoChat~\cite{li2024videochat} has two streams: a video embedder and a perception toolkit, which invokes multiple models for captions, tags, subtitles, etc. These two streams are combined and input into the final LLM with the question to produce an answer. \jian{maybe summarize video-llama and videochat as two streams methods, and give a more compact description} 
On the other hand, Video-ChatGPT~\cite{videochatgpt} leverages a single stream where the architecture first encodes each frame and then has a spatial and temporal pooling process that is finally mapped to an LLM with a linear layer. Video LLaVA~\cite{video-llava} takes advantage of the LanguageBind module to map both image and video inputs to the same embedding space. Otter~\cite{li2023otter} proposed an instruction-tuned version of OpenFlamingo~\cite{openflamingo}, such that it can also process multiple video frames as input. 
\section{MiniGPT4-Video}
\subsection{Methodology}
\begin{figure}
    \centering
    \includegraphics[width=1\linewidth]{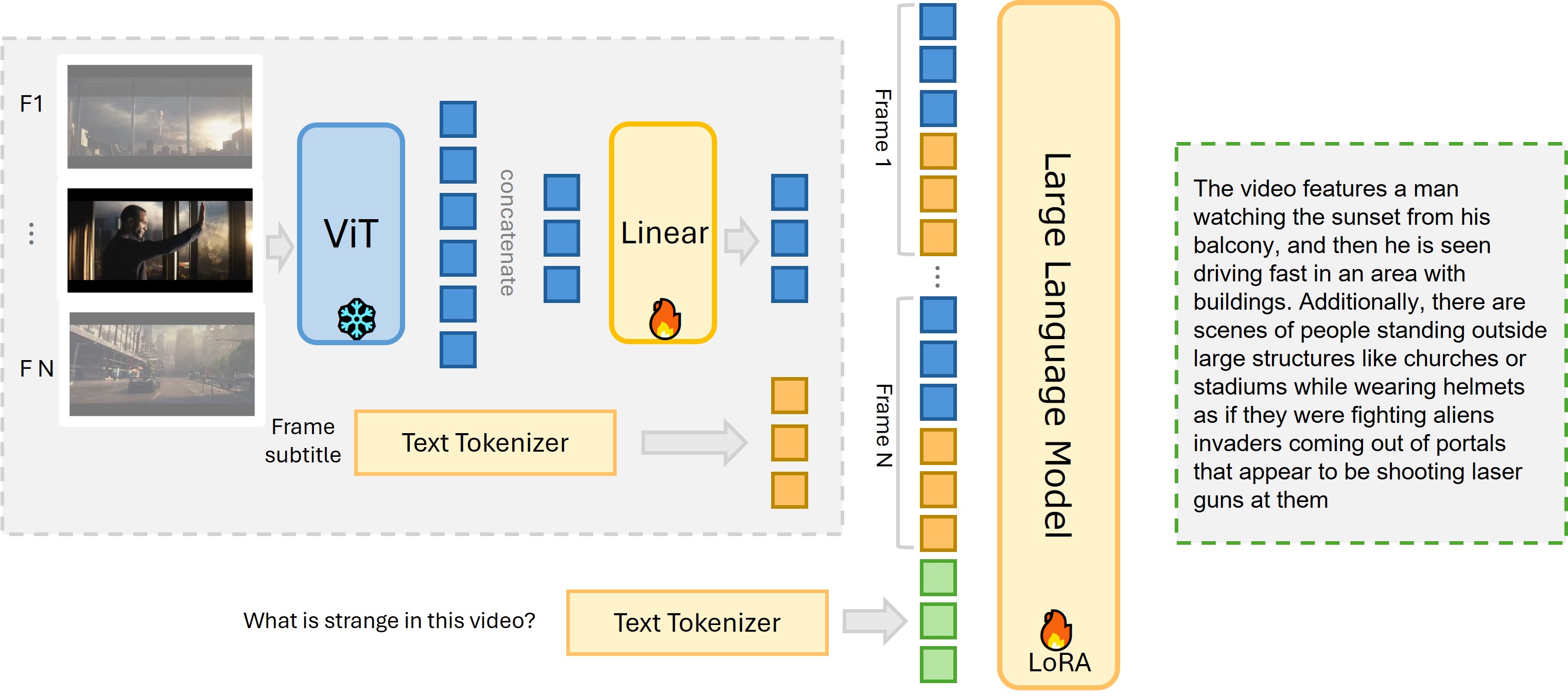}
    \caption{MiniGPT4-video architecture: For each frame, we use EVA-CLIP to get the visual tokens and concatenate each adjacent visual token into a singular token then convert these tokens to the language model space using a linear layer and get the language token from LLM tokenizer.
    Concatenate both the visual and subtitle text tokens together and do this for all the sampled frames and appending the instruction tokens at the end of the input sequence.}
    % \jian{too many typos here}}
    \label{fig:main_fig}
\end{figure}

MiniGPT-v2 \cite{minigptv2}, has successfully translated visual features into the LLM space, enabling understanding of single images. However, extending this capability to multiple frames for video comprehension entails fine-tuning the LLM to process these frames and learn the temporal dynamics.As shown in Figure \ref{fig:main_fig} Due to constraints imposed by the LLM's context window,
% , we can only sample a limited number of frames N.
% For instance, Llama 2\cite{llama2} allows sampling of up to 45 frames, whereas Mistral, by increasing the context window size, permits sampling up to 90 frames. 
% \jian{Here, we do not need to mention the details of number of frames (\eg 45, 90), and the LLM that we use, Llama 2\cite{llama2}, Mistral\cite{jiang2023mistral}. We should describe our model as a general method. These details should be moved to the implementation details. }
each video undergoes frame sub-sampling, with the number of frames (N) determined by the LLM's context window.
% ; 45 for Llama 2 \cite{llama2}and 90 for Mistral.
 Subsequently, the visual frames are aligned with textual descriptions using a pre-trained model, EVA-CLIP \cite{sun2023eva}, followed by a mapping into the large language model space using a linear layer. Similar to MiniGPT-v2~\cite{minigptv2}, we condense every four adjacent visual tokens in each image into a single token, thereby reducing token count per image by 75\%, from 256 to 64. During training the subtitles are provided with the dataset but while inference or when there is no subtitle for the video, we utilize speech-to-text model such as whisper to generate the subtitles of the video. Frame subtitles are tokenized using the LLM tokenizer, and the visual and text tokens are concatenated for each sampled frame. Instruction tokens are appended to the end of the input sequence, and the model then outputs the answer to the question.

% Efficient fine-tuning of the language model is performed using LoRA \cite{hu2021lora} (Low-Rank Adaptation).

\subsection{Training Pipeline}
\label{sec:training_pipeline}
\noindent\textbf{Large-scale image-text pair pretraining.} In the first stage, we train a linear layer, similar as~\cite{zhu2023minigpt}, which projects the visual feature encoded by the vision encoder (\eg EVA-CLIP~\cite{sun2023eva}) to the LLM's text space with captioning loss. We leverage a combined image captioning dataset that includes images from LAION~\cite{schuhmann2021laion400m}, Conceptual Captions~\cite{ sharma2018conceptual}, and SBU~\cite{ordonez2011im2text} to align the visual feature with LLM's input space. 
% To efficiently utilize the context length of LLM, we concatenate every four neighboring visual tokens into a single token, reducing the number of tokens per image by 75\% from 256 to 64.\jian{This is repeated with the previous description.}

\noindent\textbf{Large-scale video-text pair pretraining.} In the second stage, we enable the model to understand videos
% \jian{Do we need to mention short videos in this workshop paper? We do not have long video. Maybe just say video. Check this through the whole paper} 
by taking multiple frames as input. Specifically, we sample a maximum of N frames from each video. During this stage, we use the predefined prompts in the following template:\\
% \textit{\#\#\#Human: <Img><FrameFeature\_1><Sub><Subtitle text\_1>... <Img> <FrameFeature\_N><Sub><Subtitle text\_N><Instruction>\#\#\#Assistant: }\\\\
\textit{<s>[INST]<Img><FrameFeature\_1><Sub><Subtitle text\_1>... <Img> <FrameFeature\_N><Sub><Subtitle text\_N><Instruction><\//INST>}\\
 % \essam{put 90 here since Mistral takes 90 frames?}
 The number of sampled frames is contingent upon the context window of each language model,Specifically, for Llama 2\cite{llama2}, the context window is 4096 tokens, and Mistral\cite{jiang2023mistral} context window is 8192 tokens.In our approach, we represent each image with 64 tokens. Thus, for Llama 2, we designate N=45 frames, equating to 2880 tokens for visual content representation. Furthermore, we allocate 1000 tokens for subtitles, while the remaining tokens are earmarked for model output. Analogously, in the case of Mistral, where the context window is doubled, N is doubled accordingly to N=90 frames, ensuring compatibility with the extended context window.
 % where $N\leq 45$ for Llama 2 or $N\leq 90$ for Mistral.
 In this prompt, each \textit{<FrameFeature>} is replaced by the sampled video frame encoded by the visual backbone. The \textit{<Subtitle text>} represents the subtitle for the corresponding frame if applicable, and \textit{<Instruction>} represents a randomly sampled instruction from our predefined instruction set containing variant forms of instruction, such as \textit{``Briefly describe these video''}. We use combined video captioning data incorporating CMD~\cite{bain2020condensed} and WebVid~\cite{webvid} for large-scale video captioning training.

\noindent\textbf{Video question answering instruction finetuning.} In this phase, we adopt the same training strategy implemented in the second stage but focus on leveraging high-quality video-question-answering datasets for instruction fine-tuning. This fine-tuning stage helps to enhance the model's ability to interpret the input video and generate precise responses to the corresponding questions.
The template is the same as the second stage with \textit{<Instruction>} replaced by general questions as mentioned in the Video-ChatGPT~\cite{videochatgpt} dataset.

% ---------------------------------------------
% ---------------------------------------------
\subsection{Implementation Details}
\label{section:implementation_details}
% Our models are trained with 4 A100 GPUs.
 % The training process involved three distinct stages, with specific durations allocated to each. The initial stage focused on image-text training and spanned a period of two days. Subsequently, the second stage, dedicated to pre-training with video captions datasets, lasted one day, followed by the third stage, involving instruction tuning, which extended over three days. 

Throughout three training stages, we maintained a batch size of 4 and utilized the AdamW optimizer in conjunction with a cosine learning rate scheduler, setting the learning rate to 1e-4. Our visual backbone is EVA-CLIP ~\cite{sun2023eva}, with the frozen weights. Notably, we trained the linear projection layer and performed efficient fine-tuning of the language model using LoRA~\cite{hu2021lora}. Specifically, we fine-tuned the $W_q$ and $W_v$ components with a rank (\textit{r}) of 64 and a LoRA-alpha value equal 16. The entire model was trained with a consistent image resolution of $224\times224$ pixels, ensuring uniformity across all stages. 
\section{Experiments}

% -------------------------------------------------------
% \tabcolsep=0.1cm
\begin{table}[t!]
%\vspace{-8mm}
% \captionsetup{font=scriptsize}
\centering
\caption{Quantitative results on Video-ChatGPT benchmark.}
% \scalebox{0.6}{
\resizebox{0.49\textwidth}{!}{
\begin{tabular}{lcccccccc}
    \toprule
    \multicolumn{1}{c}{\multirow{3}{*}{\textbf{Method}}} & \- & \multicolumn{1}{c}{\multirow{3}{*}{\makecell{\textbf{Using} \\ \textbf{Subtitles}}}} & \- &  \multicolumn{5}{c}{\textbf{Video ChatGPT}}\\
    %\cmidrule(r){1-2}
    \cmidrule(r){5-9}
    \- & \- & \- & \- & \multirow{2}{*}{\makecell{Information\\Correctness}} & \multirow{2}{*}{\makecell{Detailed\\Orientation}}  &  \multirow{2}{*}{\makecell{Contextual\\Understanding}}  &  \multirow{2}{*}{\makecell{Temporal\\Understanding}} &  \multirow{2}{*}{Consistency}\\
    \- & \- & \- & \- & \-  & \-  &  \-  &   \-  &  \- \\

    \cmidrule(r){1-9}
    LLaMA Adapter~\cite{llama-adapter}    &\- & \xmark & \- & 2.03  & 2.32 & 2.30 & 1.98  &  2.15\\
    Video LLaMA~\cite{video-llama}  & \- & \xmark & \- &  1.96 & 2.18 & 2.16 & 1.82  & 1.79 \\
    Video Chat ~\cite{li2024videochat}  &\- & \xmark & \- & 2.23  & 2.50 & 2.53 & 1.94  & 2.24 \\
    Video-ChatGPT~\cite{videochatgpt}  &\- & \xmark & \- &  2.40 & 2.52 & 2.62 &  1.98 & 2.37 \\
    BT-Adapter-7B ~\cite{liu2023all}   & \- & \xmark & \- &  2.68 & 2.69 & 3.27 & 2.34  & 2.46 \\ 
    LLaMA-VID-7B~\cite{llamavid}     & \- & \xmark & \- &  2.96 & \textbf{3.00 }& \textbf{3.53} &  2.46 & 2.51 \\
     \rowcolor{blue!15} \textbf{Ours Llama 2-7B}   &\- & \xmark & \- &  2.93 & 2.97 & 3.45 & \textbf{2.47}  & \textbf{2.60} \\
     \rowcolor{blue!15} \textbf{Ours Mistral-7B}   &\- & \xmark & \- &  \textbf{2.97} & 2.58 &3.17 & 2.38 & 2.44\\
    
    \hline
    \rowcolor{blue!15} \textbf{Ours Llama 2-7B}   &\- & \cmark & \- &  \textbf{3.08} & \textbf{3.02} &\textbf{ 3.57} &  \textbf{2.65} & \textbf{2.67} \\
    \rowcolor{blue!15} \textbf{Ours Mistral-7B}   &\- & \cmark & \- &  3.05& 2.61&3.24& 2.42&2.42\\

  \bottomrule
\end{tabular}
% }
}
%\vspace{-3mm}
\label{tab:benchmark_videochatgpt}
\end{table}
% -------------------------------------------------------

\subsection{Datasets}
\paragraph{Training Datasets}
The Condensed Movies Video Captions dataset (CMD)\cite{bain2020condensed} comprises approximately 15,938 videos, each spanning one to two minutes in length. 
However, CMD's captions exhibit limited quality, characterized by an average sentence length of 14 words. The Webvid dataset\cite{webvid} boasts a vast collection of two million videos. 
To align with CMD's duration criteria, we refined this dataset to include videos ranging from one to two minutes in length. On the other hand, the Video Instruction Dataset~\cite{videochatgpt} offers a rich resource of 100,000 question-answer pairs distributed across 13,224 videos, distinguished by meticulous annotations. 
Noteworthy for its high-quality annotations, this dataset presents detailed answers to questions, averaging 57 words per sentence. Spanning diverse question types, including Video Summarization and Description-based QAs, it addresses spatial, temporal, relationship, and reasoning aspects, alongside creative or generative QAs.

% --------------------------------------------
\noindent\textbf{Evaluation Benchmarks}
The Video-ChatGPT benchmark~\cite{videochatgpt}, leveraging the ActivityNet-200 dataset~\cite{caba2015activitynet}, is meticulously designed to evaluate video-based conversation models' text generation capabilities across five crucial dimensions: Correctness of Information, Detail Orientation, Contextual Understanding, Temporal Understanding, and Consistency. In assessing model performance on open-ended questions, established datasets such as MSRVTT-QA~\cite{xu2017video}, MSVD-QA~\cite{xu2017video}, TGIF-FrameQA~\cite{jang2017tgifqa}, and ActivityNet-QA~\cite{yu2019activitynetqa} are employed. Furthermore, for multi-choice questions, model performance is scrutinized using the TVQA dataset~\cite{lei2019tvqa}, which is centered around popular TV shows. The validation set comprises 15,253 QA pairs, providing a robust framework for evaluation.

\subsection{Evaluation Metrics}
Aligned with the evaluation methodology established by Video-ChatGPT~\cite{videochatgpt}, we employed GPT-3.5 turbo to juxtapose model outputs with ground truth data, subsequently computing both accuracy and a score. The accuracy metric indicates the degree of correspondence between the model's output and the ground truth, while the score ranges from 0 to 5, signifying the level of alignment between the model output and the ground truth. A score of 0 indicates a significant deviation from the ground truth, while a score of 5 suggests close alignment. To ensure a fair and consistent comparison with the results presented in Video-ChatGPT~\cite{videochatgpt}, we adopted the same prompt for our evaluations.

\subsection{Results}
For a comprehensive evaluation of our proposed architecture, we assessed its performance across three benchmark types: Video-ChatGPT, Open-ended Questions, and Multiple-Choice Questions (MCQs).
In the Video-ChatGPT benchmark, depicted in Table \ref{tab:benchmark_videochatgpt}, our model is comparable with the previous methods without subtitles. When we add the subtitles as input, our model achieves the state-of-the-art in all five dimensions, which verified that our model can utilize the subtitle information to improve the video understanding. 
In the zero-shot evaluation of open-ended and multiple-choice question benchmarks, as illustrated in Figure \ref{fig_teaser} and Table~\ref{tab:zeroshot_evaluation}, our proposed MiniGPT4-Video significantly outperforms existing state-of-the-art methods. It achieves notable margins of improvement 4.22\%, 1.13\%, 20.82\%, and 13.1\% on the MSVD, MSRVTT, TGIF, and TVQA benchmarks, respectively.
% For open-ended questions, we utilized four datasets: MSVD, MSRVTT, TGIF, and ActivityNet. These datasets serve as valuable resources for assessing model performance on diverse question types.
% In the case of MCQs, we employed the TVQA dataset, which features vision questions answerable solely from visual cues, as well as text questions answerable from subtitles alone. 
% \jian{"we evaluated models based on five crucial aspects ..., in the case of MCQs". These descriptions duplicate those in Evaluation Benchmarks. And it seems that we did not have any number or performance analyses for table 1.} 
The results, both with and without subtitles as shown in Table \ref{tab:zeroshot_evaluation}, further demonstrate that integrating subtitle information alongside visual cues significantly enhances performance, with accuracy rising from 33.9\% to 54.21\% on TVQA. While subtitles contribute substantially to performance improvements on TVQA, their inclusion doesn't offer added value for datasets like MSVD-QA, MSRVTT-QA, TGIF-QA, and ActivityNet, where questions are exclusively vision-based.

% \jian{why mention table 1 here? all the mentioned datasets are not in table 1.}

% -------------------------------------------------------
\tabcolsep=0.1cm
\begin{table}[t!]
% \captionsetup{font=small}
\centering
\caption{Zeroshot evaluation for open ended question and Multiple choices questions on MSVD, MSRVTT, TGIF, ActivityNet and TVQA.NA indicates not applicable, while MSVD and TGIF videos do not have audio.}
% \scalebox{0.65}{
\resizebox{0.49\textwidth}{!}{
\begin{tabular}{lccccccccccccccccc}
    \toprule
    \multicolumn{1}{c}{\multirow{3}{*}{\textbf{Method}}} & \- & \multicolumn{1}{c}{\multirow{3}{*}{\makecell{\textbf{Using} \\ \textbf{Subtitles}}}} & \- &  \multicolumn{8}{c}{\textbf{Open Ended Questions}} & \- &\multicolumn{1}{c}{\textbf{MCQ}}\\
    %\cmidrule(r){1-2}
    \cmidrule(r){5-12}
    \cmidrule(r){14-14}
    \- & \- & \- & \- & \multicolumn{2}{c}{MSVD} &\multicolumn{2}{c}{MSRVTT} &\multicolumn{2}{c}{TGIF} &\multicolumn{2}{c}{ActivityNet}  & \- &\multicolumn{1}{c}{TVQA}\\
    %\cmidrule(r){1-2}
    \cmidrule(r){5-6}
    \cmidrule(r){7-8}
    \cmidrule(r){9-10}
    \cmidrule(r){11-12}
    \cmidrule(r){14-14}
    \- & \- & \- & \- & \- Acc.↑ & Score↑ & Acc.↑ & Score↑ &  Acc.↑ & Score↑ &  Acc.↑ & Score↑ & \- &  Acc.↑ \\
    \cmidrule(r){1-4}
    \cmidrule(r){5-6}
    \cmidrule(r){7-8}
    \cmidrule(r){9-10}
    \cmidrule(r){11-12}
    \cmidrule(r){14-14}
    FrozenBiLM \cite{yang2022zeroshot}       &\- & \xmark & \- & 32.2  & --  & 16.8& -- & 41  & -- & 24.7  & -- & \- & 29.7 \\
    LLaMA Adapter~\cite{llama-adapter}    &\- & \xmark & \- & 54.9  & 3.1  & 43.8&2.7  & --  & -- & 34.2  & 2.7 &  \- &  --\\
    Video LLaMA~\cite{video-llama}  & \- & \xmark & \- & 51.6  & 2.5  & 29& 1.8 & --  &  --& 12.4  & 1.1 &  \- & -- \\
    Video Chat ~\cite{li2024videochat}  &\- & \xmark & \- & 56.3  &  2.8 & 45& 2.5 & 34.4  & 2.3 & 26.5  & 2.2 &  \- & -- \\
    
    Video-ChatGPT~\cite{videochatgpt}  &\- & \xmark & \- &  64.9 & 3.3  & 49.3& 2.8 & 51.4  & 3.0 & 35.2  & 2.7 & \- & 23.35 \\
    BT-Adapter-7B ~\cite{liu2023all}   & \- & \xmark & \- & 67.7  & 3.7  & 57& 3.2 & --  & -- & 45.7  & 3.2 & \- &  --\\ 
    LLaMA-VID-7B~\cite{llamavid}     & \- & \xmark & \- &  69.7 & 3.7  & 57.7& 3.2 & --  & -- & \textbf{ 47.4} & 3.3 &  \- & --  \\
    %LLaMA-VID-13B~\cite{llamavid} & \- & \xmark & \- &  70 & 3.7  & 58.9& 3.3 & --  & -- & 47.5  & 3.3 &  \- &  \\
    %Video-LLaVA-13B~\cite{video-llava}    & \- & \xmark & \- &  70.7 &  3.9 & 59.2& 3.5 & 70  & 4.0&  45.3 & 3.3 &  \- &  36.58\\
    \rowcolor{blue!15} \textbf{Ours Llama 2-7B}   &\- & \xmark & \- &  72.93 & 3.84  & \textbf{58.83}& \textbf{3.29} &  67.9 &3.71 & 45.85  & 3.23 &  \- &  \textbf{36.45}\\
    \rowcolor{blue!15} \textbf{Ours Mistral-7B}   &\- & \xmark & \- &  \textbf{73.92} & \textbf{4.06}  & 58.26& 3.52 &  \textbf{72.22} &\textbf{4.08}&  44.25 &3.35  &  \- & 33.9 \\
    \hline
    
    \rowcolor{blue!15} \textbf{Ours Llama 2-7B}   &\- & \cmark & \- &  N/A & N/A  & \textbf{59.73}& \textbf{3.3} & N/A & N/A  &  46.3 & \textbf{3.4} &  \- &  46.94\\
    \rowcolor{blue!15} \textbf{Ours Mistral-7B}   &\- & \cmark & \- &  N/A & N/A  & 58.68& 3.53 & N/A &N/A&  44.38& 3.36 &  \- &  \textbf{54.21}\\
  \bottomrule
\end{tabular}
}
\label{tab:zeroshot_evaluation}
\end{table}
% -------------------------------------------------------

\section{Qualitative Results}
\label{section:qual}

Here in this section we show some qualitative results for our model to show the performance of it and its ability to answer different questions.
For each example you can open the video link which attached to the figure description to watch the video.

\begin{figure}[t!]
    \centering
    \includegraphics[width=1.0\linewidth]{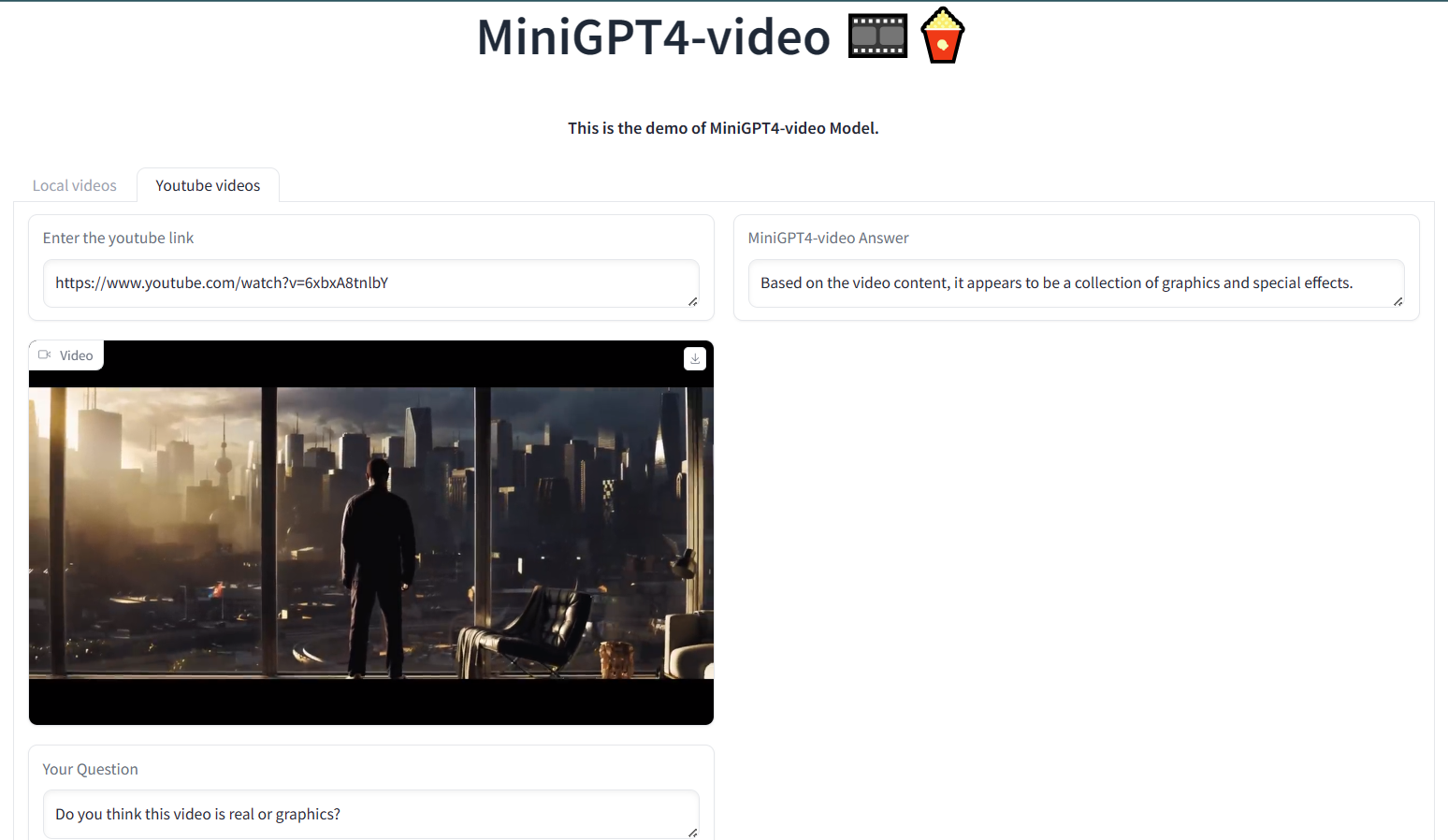}
    \captionsetup{font=scriptsize}
    \caption{Qualitative result of MiniGPT4-video for understanding general videos \href{https://www.youtube.com/watch?v=6xbxA8tnlbY&ab_channel=UnrealEngine}{[link]}.}
    \label{fig:short1}
\end{figure}

% \begin{figure}[t!]
%     \centering
%     \includegraphics[width=1.0\linewidth]{figures/4.PNG}
%     \captionsetup{font=scriptsize}
%     \caption{Qualitative result of MiniGPT4-video  understanding catron videos  \href{https://www.youtube.com/watch?v=q99eI\_uDDGc\&ab\_channel=WarnerBros.Entertainment}{[link]}.}
%     \label{fig:short2}
% \end{figure}

\begin{figure}[t!]
    \centering
    \includegraphics[width=1.0\linewidth]{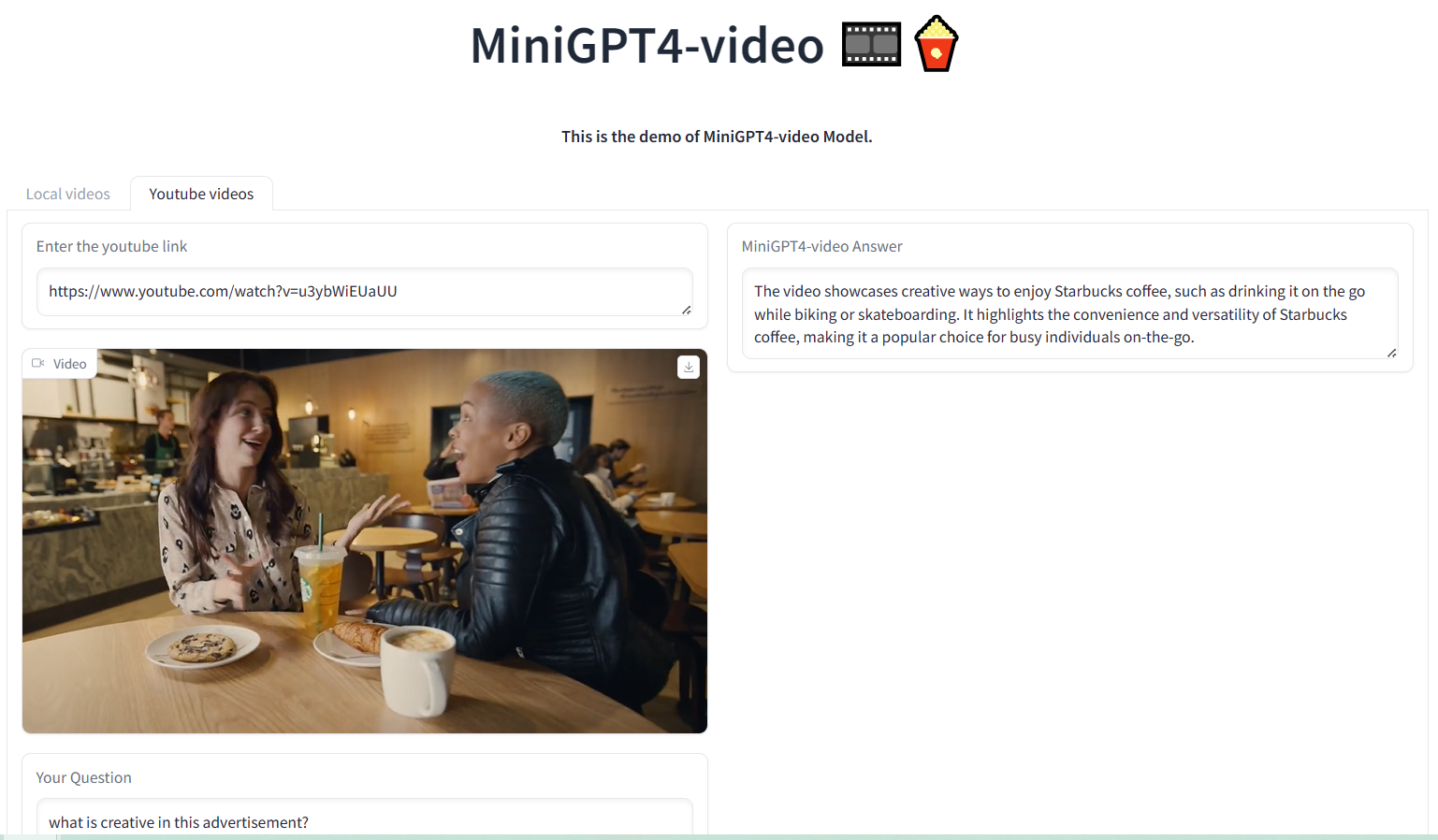}
    \captionsetup{font=scriptsize}
    \caption{Qualitative result of MiniGPT4-video  \href{https://www.youtube.com/watch?v=u3ybWiEUaUU}{[link]}.}
    \label{fig:short2}
\end{figure}

\begin{figure}[t!]
    \centering
    \includegraphics[width=1.0\linewidth]{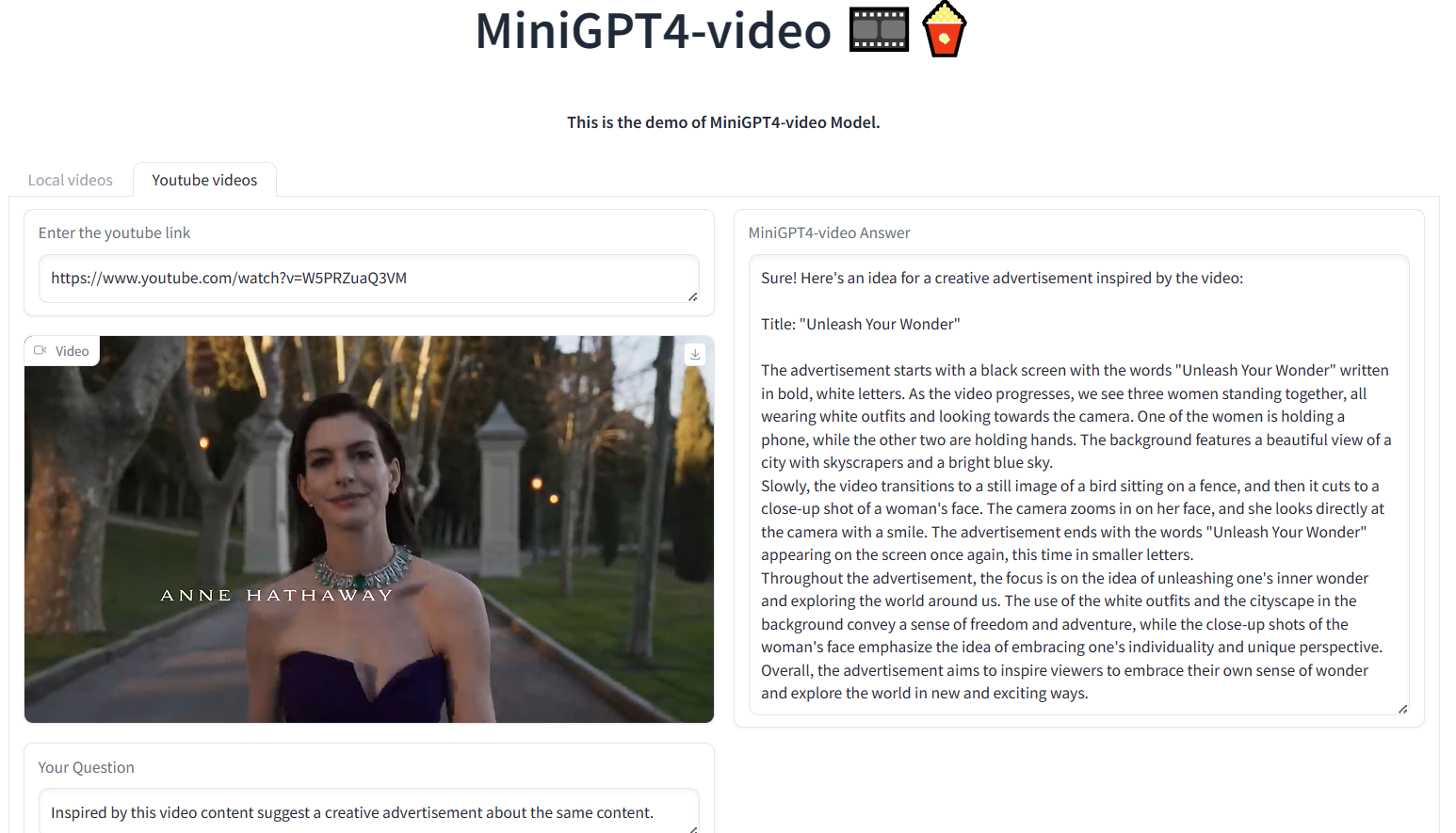}
    \captionsetup{font=scriptsize}
    \caption{Qualitative result of MiniGPT4-video  \href{https://www.youtube.com/watch?v=W5PRZuaQ3VM}{[link]}.}
    \label{fig:short2}
\end{figure}

\begin{figure}[t!]
    \centering
    \includegraphics[width=1.0\linewidth]{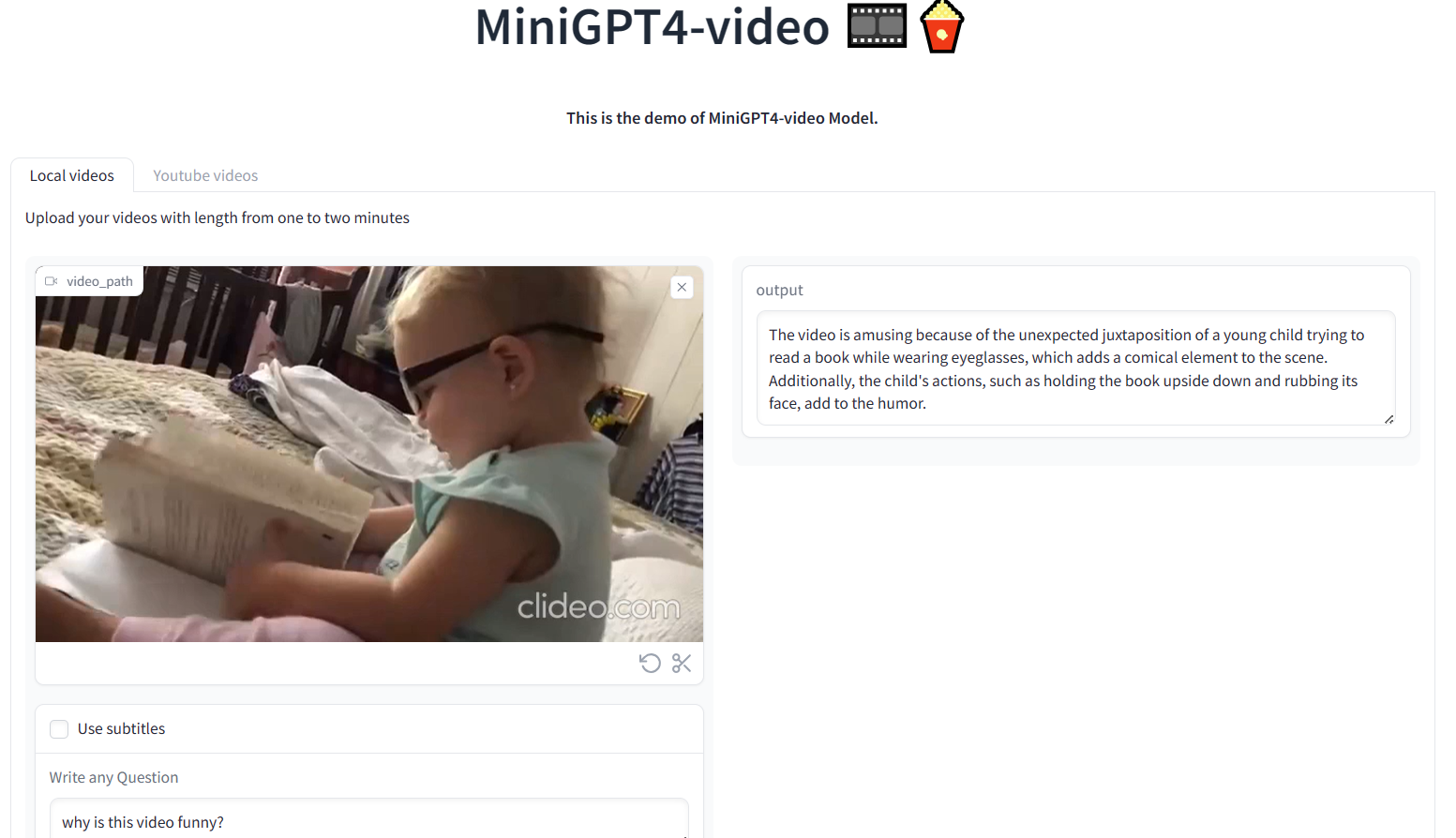}
    \captionsetup{font=scriptsize}
    \caption{Qualitative result of MiniGPT4-video}
    \label{fig:short2}
\end{figure}

\begin{figure}[t!]
    \centering
    \includegraphics[width=1.0\linewidth]{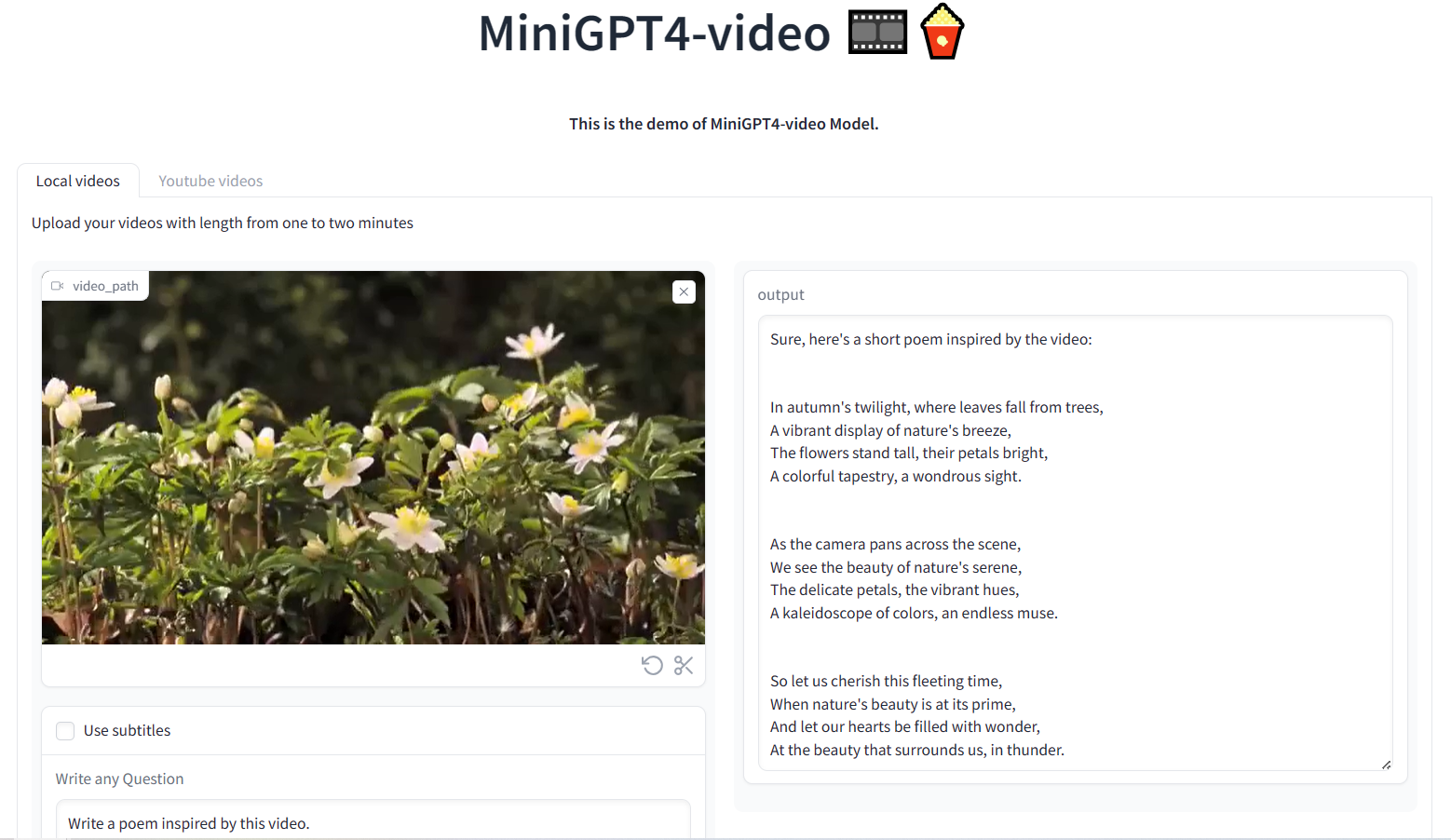}
    \captionsetup{font=scriptsize}
    \caption{Qualitative result of MiniGPT4-video}
    \label{fig:short2}
\end{figure}

\begin{figure}[t!]
    \centering
    \includegraphics[width=1.0\linewidth]{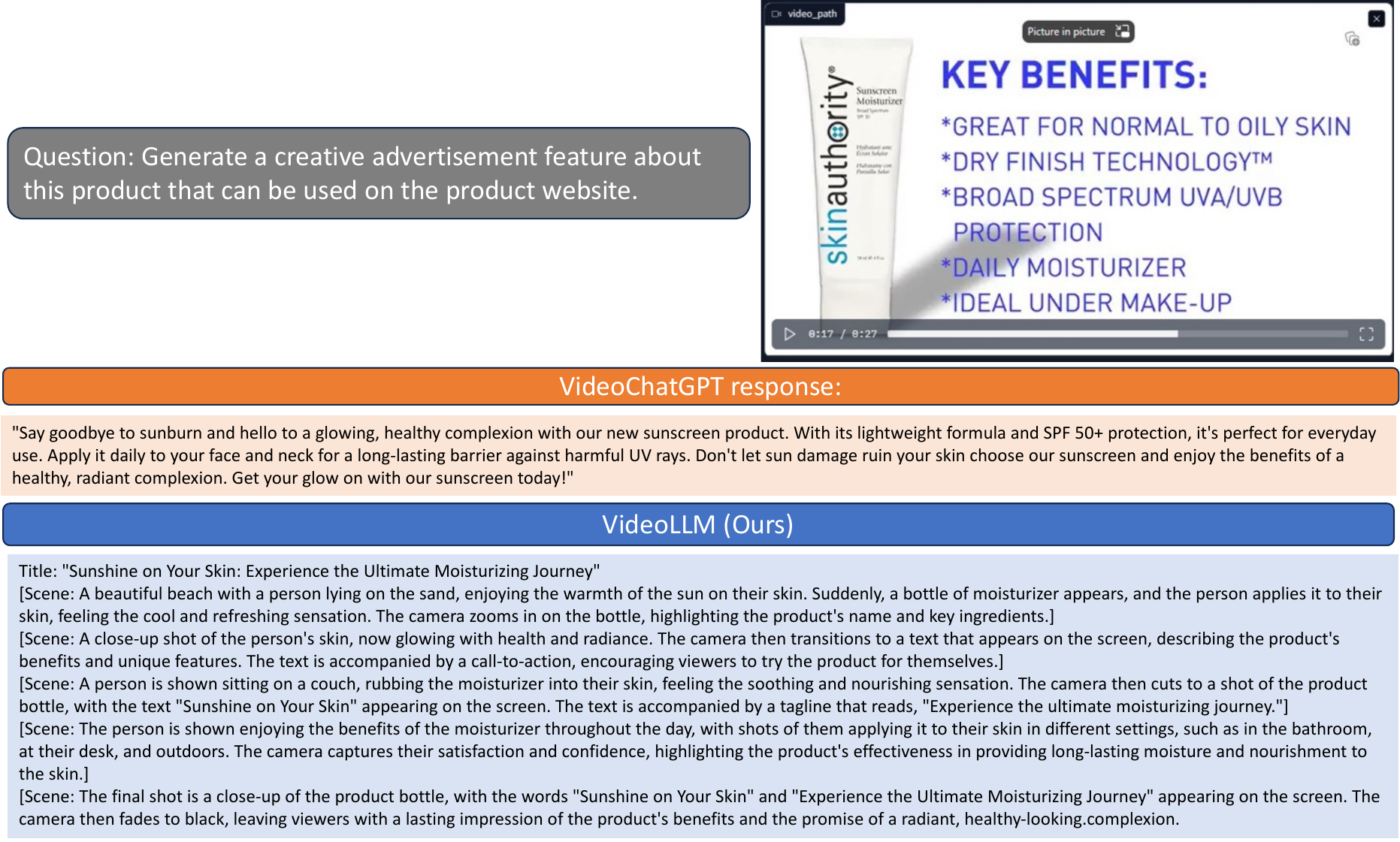}
    \captionsetup{font=scriptsize}
    \caption{\textbf{MiniGPT4-video vs VideoChatGPT~\cite{videochatgpt}.}}
    \label{fig:visionllm}
\end{figure}

\begin{figure}[t!]
    \centering
    \includegraphics[width=1.0\linewidth]{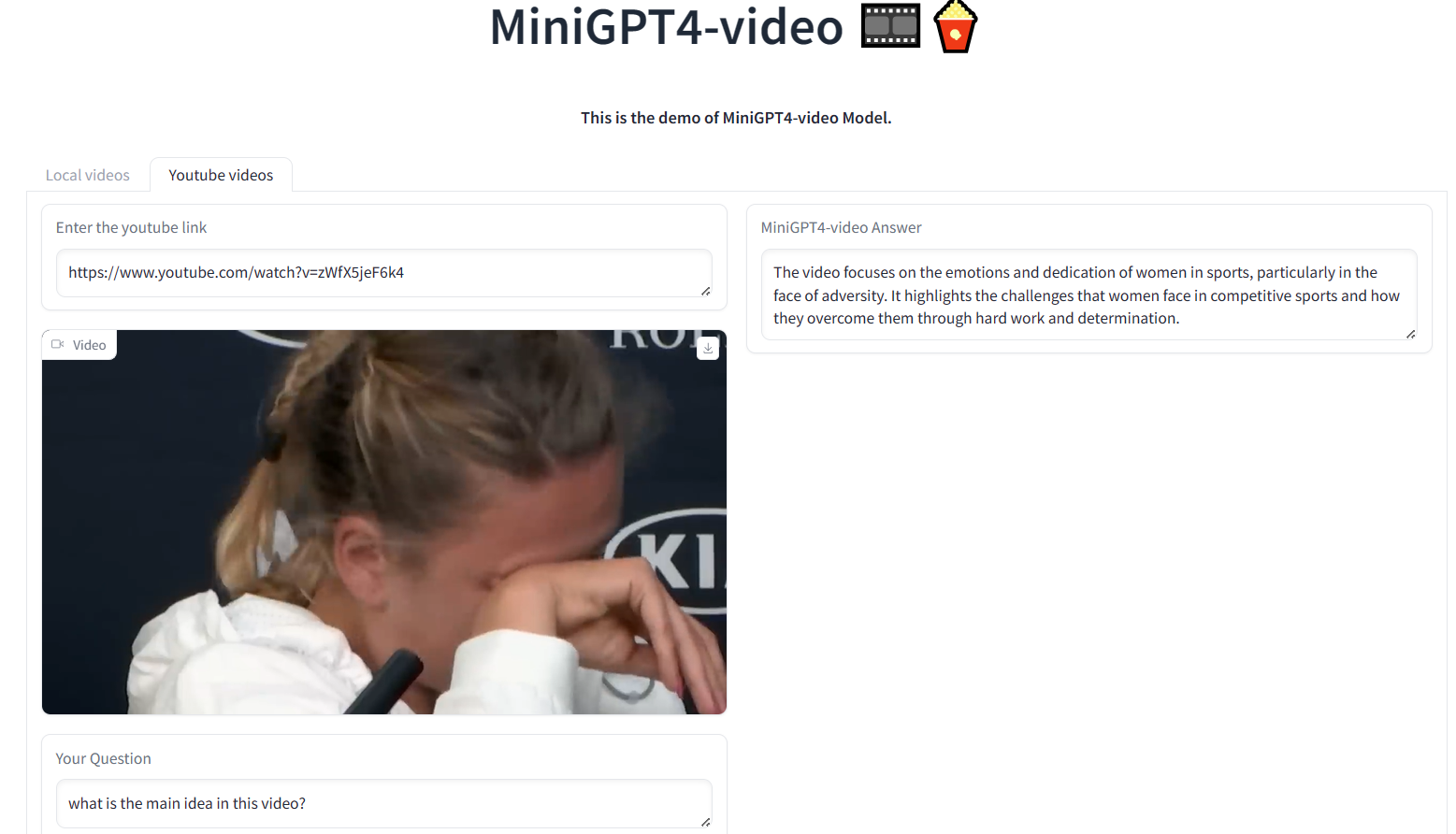}
    \captionsetup{font=scriptsize}
    \caption{Qualitative result of MiniGPT4-video \href{https://www.youtube.com/watch?v=zWfX5jeF6k4}{[link]}.}
    \label{fig:visionllm}
\end{figure}

\section{Conclusion}

In summary, MiniGPT4-Video offers a compelling solution for video question answering, effectively amalgamating visual and conversational comprehension within the video domain. By directly inputting both visual and textual tokens, MiniGPT4-Video empowers the Language Modeling Model (LLM) to grasp the intricate relationships between video frames, showcasing promising proficiency in understanding temporal dynamics within video content.
Despite its notable achievements, MiniGPT4-Video faces a limitation imposed by the context window of the LLM. Specifically, the current version requires video lengths of 45 frames for the Llama 2 version (equivalent to less than one and a half minutes at a sampling rate of 0.5 frames per second) and 90 frames for the Mistral version (equivalent to less than three minutes). Future research endeavors will focus on extending the model's capabilities to handle longer video sequences, thereby addressing this limitation and further enhancing its applicability and effectiveness in real-world scenarios.
% \clearpage
{
    \small
    \bibliographystyle{ieeenat_fullname}
    \bibliography{main}
}


\end{document}